\documentclass[9pt]{article}
\usepackage{amsmath,graphicx, multicol}

\def\x{{\mathbf x}}

\usepackage{xcolor}

\usepackage{hyperref}

\usepackage{amsmath}
\usepackage{amsthm}
\usepackage{amssymb}
\usepackage[nice]{nicefrac}
\newcommand{\h}{\mathbf{h}}
\newcommand{\w}{\mathbf{w}}
\newcommand{\G}{\mathbf{G}}
\newcommand{\W}{\mathbf{W}}
\newcommand{\M}{\mathbf{M}}
\newcommand{\mub}{\boldsymbol{\mu}}
\newcommand{\thetab}{\boldsymbol{\theta}}

\usepackage[a4paper,margin = 1in]{geometry}
\newenvironment{Figure}
  {\par\medskip\noindent\minipage{\linewidth}}
  {\endminipage\par\medskip}
\usepackage{caption}
\usepackage{titlesec}
\titleformat*{\section}{\large\bfseries}
\title{Engineering the Neural Collapse Geometry of Supervised-Contrastive Loss}
\author{Jaidev Gill \qquad Vala Vakilian \qquad Christos Thrampoulidis \footnote{JG and CT gratefully acknowledge the support of an NSERC USRA. The authors also acknowledge use of the Sockeye cluster by UBC Advanced Research Computing.  This work is supported by an an NSERC Discovery Grant, NSF Grant CCF-2009030, and by a CRG8-KAUST award.} \\ University of British Columbia}

\newtheorem{remark}{Remark}
\newtheorem{proposition}{Proposition}

\theoremstyle{definition}
\newtheorem{definition}{Definition}

\begin{document}

\maketitle
\begin{multicols}{2}

\begin{abstract}
 
Supervised-contrastive loss (SCL) is an alternative to cross-entropy (CE) for classification tasks that makes use of similarities in the embedding space to allow for richer representations. In this work, we propose methods to engineer the geometry of these learnt feature embeddings by modifying the contrastive loss. In pursuit of adjusting the geometry we explore the impact of prototypes, fixed embeddings included during training to alter the final feature geometry. Specifically, through empirical findings, we demonstrate that the inclusion of prototypes in every batch induces the geometry of the learnt embeddings to align with that of the prototypes. We gain further insights by considering a limiting scenario where the number of prototypes far outnumber the original batch size. Through this, we establish a connection to cross-entropy (CE) loss with a fixed classifier and normalized embeddings. We validate our findings by conducting a series of experiments with deep neural networks on benchmark vision datasets. 

\end{abstract}
%
%
\section{Introduction}
\label{sec:intro}

Understanding the structure of the learned features of deep neural networks has gained significant attention through a recent line of research surrounding a phenomenon known as \textit{Neural Collapse} (NC) formalized by \cite{PHD20}. 
The authors of \cite{PHD20} have found that, when training a deep-net on balanced
datasets with cross-entropy (CE) loss beyond zero training
error, the feature embeddings collapse to their corresponding
class mean and align with the learned classifier, overall forming
a simplex equiangular tight frame (ETF) geometry. In
other words, at the terminal phase of training, the class-mean
embeddings form an implicit geometry described by vectors
of equal norms and angles that are maximally separated. Following
\cite{pmlr-v206-behnia23a}, we call this geometry ``implicit,'' since it is not
enforced by an explicit regularization, but rather induced by common optimizers, such as SGD.
A number of followup studies have provided further analysis on the NC phenomenon \cite{NC_minCol, NC_Geom, NC_Mixon} and extended the implicit-geometry characterization of CE to imbalanced data \cite{seli, pmlr-v206-behnia23a}. At the same time, \cite{graf21a, BCL, kini2023supervisedcontrastive} have 
\begin{Figure}
        \centering
        \includegraphics[width = 0.98\textwidth]{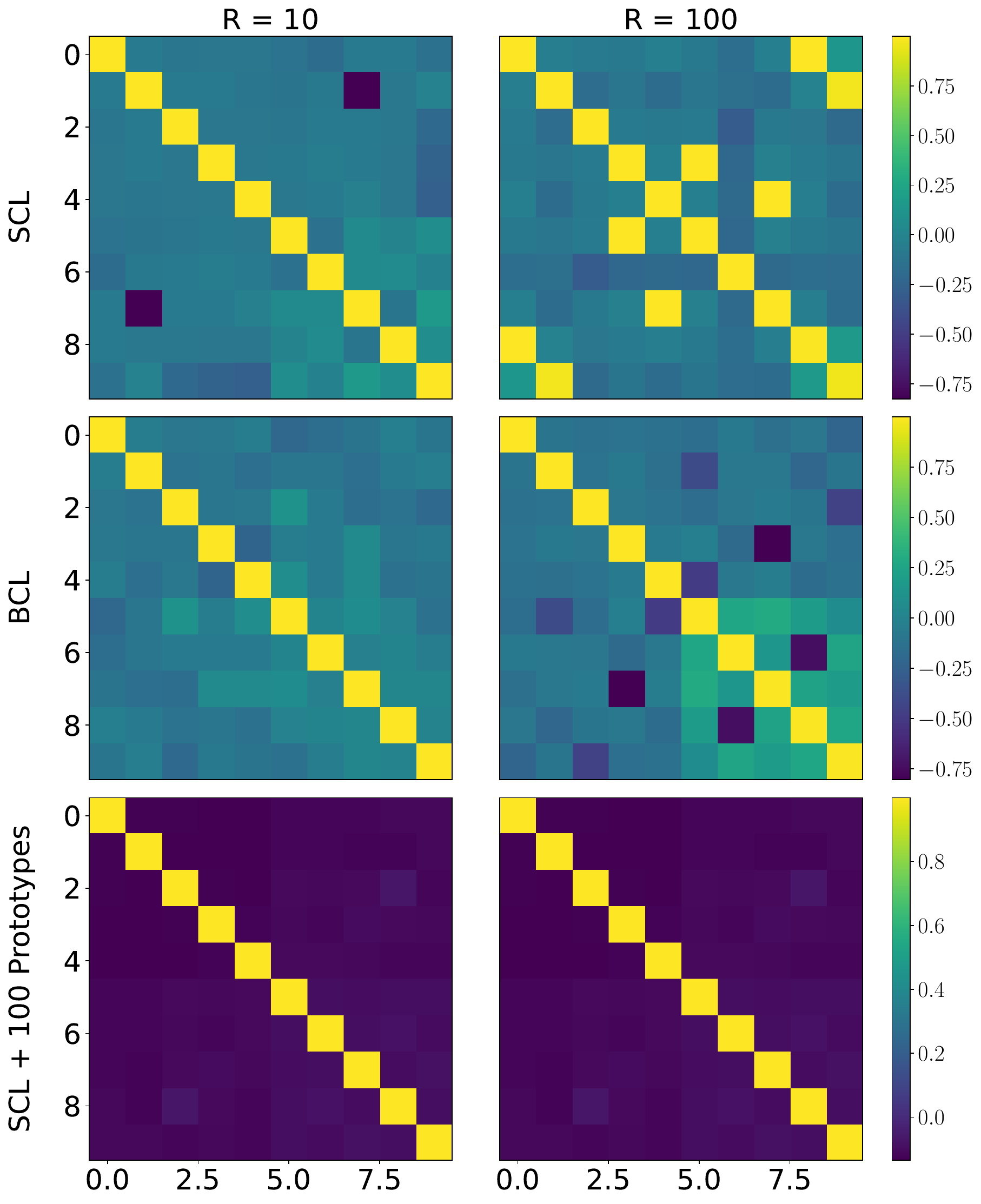}
        \captionof{figure}{Comparison of Gram matrices $\G_\M$ at last epoch (350) trained on STEP imbalanced CIFAR-10 and ResNet-18 with (Top) vanilla SCL ($n_w = 0$) (Middle) Class averaging (BCL) \cite{BCL} satisfying class representation requirements through batch binding \cite{kini2023supervisedcontrastive} (Bottom) SCL with ($n_w = 100$) prototypes.}
        \label{fig:heatmap_comp} 
\end{Figure}
\noindent shown, both empirically and theoretically, that such characterizations can be extended to other loss functions, specifically the supervised-contrastive loss (SCL).

Drawing inspiration from unsupervised contrastive learning  \cite{simCLR}, SCL was proposed by \cite{khosla_NEURIPS2020} as a substitute to CE for classification. Specifically, SCL makes use of semantic information by directly contrasting learned features. \cite{graf21a} was the first to  theoretically analyze the implicit geometry of SCL, demonstrating that it forms  an ETF when data is balanced.
However, when the label distribution is imbalanced,  the geometry changes. To combat this, \cite{BCL} proposed a training framework, which they called balanced contrastive learning (BCL), improving the generalization test accuracy under imbalances. Their framework uses a class averaging modification to SCL alongside a set of $k$ \emph{trainable} prototypes,
representing class centers, trained using a logit adjusted cross-entropy \cite{cao2019learning, BCL_longtail_Method}.

According to the authors, the BCL framework drives the implicit geometry to an ETF. In another related work, drawing inspiration from unsupervised frameworks such as MoCo \cite{MoCO}, \cite{cui2021parametric} introduced PaCo, a supervised contrastive method that also takes advantage of such trainable class centers to  improve test accuracy under imbalances. 

These works collectively suggest that prototypes can play a crucial role in determining the implicit geometry when training with SCL.  However, in their respective frameworks, prototypes are treated as trainable parameters, optimized alongside various other heuristics and modifications.  Thus, it is challenging to ascertain their specific impact on the training process. This raises the question \textit{what is the direct impact of prototypes on the SCL geometry when isolated from other modifications?} 

In order to answer this question, this paper investigates the implicit geometry of SCL with \emph{fixed} prototypes, departing from the conventional approach of using trainable prototypes. We introduce a new method to incorporate fixed prototypes to the SCL training by  augmenting each batch with copies of all class prototypes. Our experimental results demonstrate that choosing prototypes that form an ETF leads to a remarkably accurate convergence of the embeddings' implicit geometry to an ETF, \emph{regardless of imbalances}. Furthermore, this convergence is achieved with a moderate number of prototype copies per batch.  Importantly, we argue that the computational overhead incurred by our SCL approach with fixed prototypes remains independent of the number of copies of prototypes, motivating an investigation into its behavior as the number of copies increases. In this limit, we derive a simplified form of SCL and subsequently prove that the implicit geometry indeed becomes an ETF when ETF prototypes are selected. Intriguingly, this simplified SCL form resembles the CE loss with fixed classifiers, albeit utilizing normalized embeddings. Finally, realizing the flexibility of choosing prototypes that form an arbitrary target geometry, we pose the question: \emph{Is it possible to tune the learned features to form an arbritrary and possibly asymmetric  geometry?} Through experiments on deep-nets and standard datasets, we demonstrate that by selecting prototypes accordingly we can achieve implicit geometries that deviate from symmetric geometries, such as ETFs. 

\section{Tuning Geometry with Prototypes}
\label{sec:tuning}

 \noindent\textbf{Setup.} We consider a $k$-class classification task with training dataset $\mathcal{D} = \{ (\x_i, y_i) : i \in [N] \}$ where $\x_i \in \mathbb{R}^p$ are the $N$ training points with labels $y_i\in[k]$.\footnote{We denote $[N] := \{1, 2, \dots, N \}$.} The SCL loss is optimized over batches $B\subset[N]$ belonging to a batch-set $\mathcal{B}$. Concretely, $\mathcal{L} := \sum_{B \in \mathcal{B}}\mathcal{L}_B$, where the loss for each batch $B$ is given below as 
\begingroup
\fontsize{9.0pt}{10.0pt} \selectfont
\begin{align}\label{eq:loss}
    \hspace{-0.08in} \mathcal{L}_B :=\sum_{i\in B}\frac{1}{n_{B,y_i}-1}\sum_{\substack{j\in B\\j \neq i \\ y_j = y_i}}\log\big(\sum_{\substack{\ell \in B \\ \ell \neq i}}\exp{(\h_i^\top\h_\ell - \h_i^\top\h_j)}\big)\,.
\end{align}
\endgroup
 Here, $\h_i:=\h_{\thetab}(\x_i)\in\mathbb{R}^d$ is the last-layer learned feature-embedding corresponding to the original training point $\x_i$ for a network parameterized by parameters $\thetab$. Also, $n_{B,y_i}$ is the number of samples sharing the same label as $\h_i$ in the batch $B$. Lastly, we let $|B| = n$ be the batch size. As per standard practice \cite{simCLR,khosla_NEURIPS2020}, we assume a normalization layer as part of the last layer, hence 
 $\|\h_i\| = 1$ $\forall i \in [N]$. It is also common to include a scaling of the inner products by a temperature parameter $\tau$ \cite{khosla_NEURIPS2020}; since this can be absorbed in the normalization, we drop it above for simplicity.

\vspace{2pt}
\noindent\textbf{Methodology.} Inspired by  the class-complement method of \cite{BCL}, the learnable class centers of \cite{cui2021parametric}, and the batch-binding algorithm of \cite{kini2023supervisedcontrastive}, we propose using \emph{fixed} prototypes. These prototypes collectively form a desired reference geometry for the embeddings to learn. 
\begin{definition}[Prototype]
     A \textit{prototype} $\w_c \in \mathbb{R}^d$ for class $c\in [k]$ is a fixed vector that represents the desired representation of embeddings $\{\h_i\}_{y_i = c}$ in class $c$. 
\end{definition}
Our method optimizes SCL with a new batch $\{ \h_i \}_{i\in B} \cup \mathcal{W}$, where $\mathcal{W} := \bigcup_{i=1}^{n_w} \{ \w_1, \w_2 , \dots, \w_k \}$ and $n_w$ is the number of added prototypes per class. We highlight two key aspects of this strategy. (i) First, as $n_w$ increases, there is \emph{no increase} in the computational complexity of the loss computation. This is because the number of required inner product computations between embeddings increases from $\nicefrac{n^2}{2}$ in vanilla SCL (Eq. \eqref{eq:loss}) to $\nicefrac{n^2}{2} + nk$ when prototypes are introduced. This increase is solely due to the presence of $k$ distinct prototypes and remains constant regardless of the value of $n_w.$
As we will see, this aspect becomes critical as increasing the number of prototypes can help the learned embeddings converge faster to the chosen prototype geometry (see Defn.~\ref{def:geometry}) with minimal added computational overhead (at least when $k=O(n)$). (ii) Second, we guarantee that prototypes are fixed and form a suitable, engineered geometry, defined formally in Definition \ref{def:geometry} below. In particular, this is in contrast to \cite{cui2021parametric} where prototypes are learned, and \cite{BCL} which conjectures that the trained prototypes form an ETF.

\begin{definition}[Prototype Geometry] \label{def:geometry}
     Given a set of prototypes $\{\w_c\}_{c \in [k]}$ the prototype geometry is characterized by a symmetric matrix $\G_* = \W^\top\W$ where $\W = [\w_1 \cdots \w_k]$. 
\end{definition}



\begin{Figure}
        \centering
        \includegraphics[width = \textwidth]{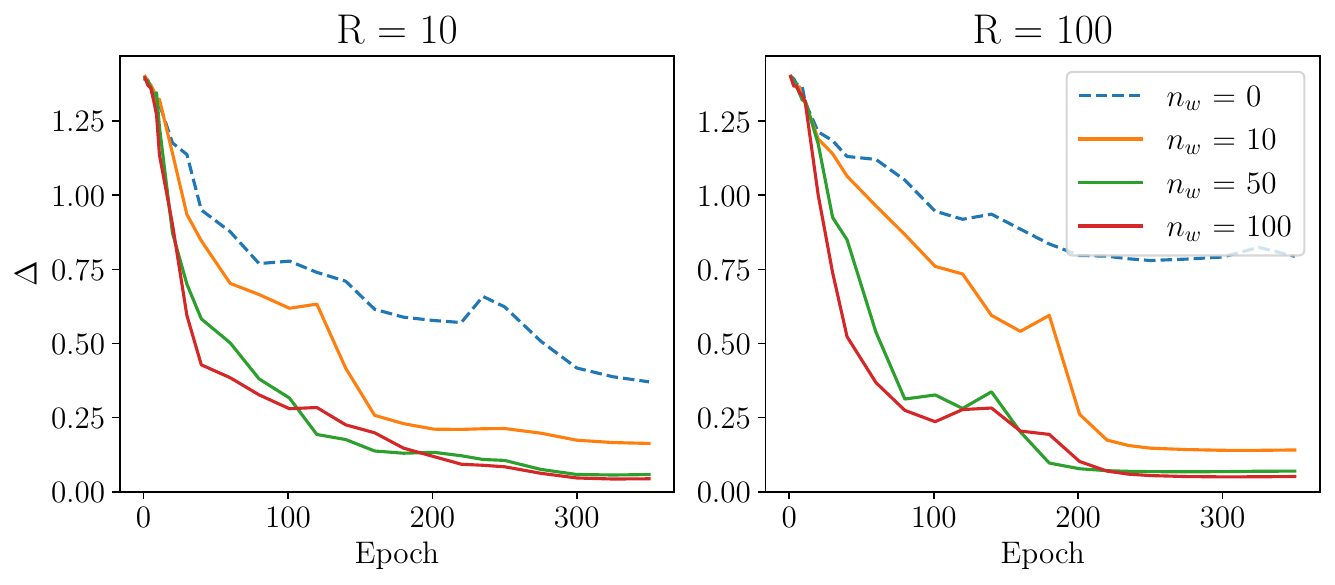}
        \captionof{figure}
        {Convergence metric $\Delta_{\G_{ETF}}$ tracked throughout training of ResNet-18 on STEP-imbalanced ($R = 10, 100$) CIFAR-10 while varying the number $n_w$ of prototypes per class. As $n_w$ increases, the feature geometry exhibits a stronger convergence to ETF.}
        \label{fig:convergence} 
\end{Figure}

\noindent\textbf{Experiments.} To display the impact of prototypes on feature geometry, we train a ResNet-18 \cite{ResNet} backbone with a two layer MLP projector head \cite{khosla_NEURIPS2020} using prototypes. We train the model with batch doubling \cite{kini2023supervisedcontrastive} resulting in a batch size of $n = 2048$, a constant learning rate of $0.1$, and temperature parameter $\tau = 0.1$ as in \cite{kini2023supervisedcontrastive}. We modify CIFAR-10 ($k = 10$) such that the first $5$ classes contain 5000 examples and the last $5$ classes have $5000/R$ samples with imbalance ratios $R=10,100$. In Fig.~\ref{fig:heatmap_comp}, we compare the final epoch geometry of models trained with vanilla SCL, BCL \cite{BCL} without prototype training and logit-adjusted CE \cite{BCL_longtail_Method}, and, SCL with 100 prototypes per class. The figure suggest that the embeddings trained with SCL and prototypes form an ETF geometry irrespective of imbalance ratio. On the other hand, SCL and BCL geometries are noticeably less symmetric with angles between minority centers decreasing with higher imbalance ratios. This highlights the impact of prototypes on achieving an ETF geometry, and further emphasizes their importance within frameworks such as BCL.

To study the impact of the number of prototypes we define the concept of Geometric Convergence (see Defn.~\ref{def:convergence} below) and compare the convergence to ETF ($\Delta_{\G_{ETF}}$) when training with SCL using different number of prototypes $n_w = 0,10,50,100$. As illustrated in Fig. \ref{fig:convergence}, without prototypes ($n_w = 0$) SCL does \emph{not} converge to ETF. However, simply adding 100 total prototype examples to the batch significantly improves convergence to the ETF geometry ($n_w = 10$). Moreover, once the prototypes make up $\sim 20\% $ of the batch size, convergence is nearly perfect (see $n_w = 50$). This observation motivates the study of SCL when $n_w$ outnumbers the training datapoints within the batch.


\begin{definition}[Geometry Convergence] \label{def:convergence}
We say that the geometry of learned embeddings has successfully converged if $\G_\M \rightarrow \G_*$, where $\G_*$ is as given in Defn.~\ref{def:geometry}. Here, $\G_\M = \M^\top\M$, $\M= [ \mub_1 \cdots  \mub_k ]$ where $\mub_c = \text{Ave}_{y_i = c} \h_i$. As a measure of convergence, we track $\Delta_{\G_*} = \|\nicefrac{\G_\M}{\|\G_\M\|_F} - \nicefrac{\G_*}{\|\G_*\|_F}\|_F$.
\end{definition}

\section{Connection to Cross-Entropy}
\label{sec:celimit}


Having seen the impact of increasing the number of prototypes ($n_w$) on the learnt geometry, it is natural to ask how the prototypes impact the loss as these prototypes begin to outnumber the original number of samples in the batch. Further, this sheds light on prototype-based methods that help improve test accuracy such as BCL \cite{BCL} and PaCO \cite{cui2021parametric} as both losses include multiplicative hyperparameters to tune the impact of prototypes.

\begin{proposition} \label{thm: ce_limit}
Let $\hat{n} := k \cdot n_w$ be the total number of prototypes added to the batch, and $n$ be the original batch size. Then in the limit $\hat{n} \gg n$ the batch-wise SCL loss becomes, 
\begingroup
\fontsize{9.0pt}{10.0pt} \selectfont
\begin{equation*} \mathcal{L}_{B} \rightarrow
    - \sum_{i \in B} \left[ \log \left( \frac{\exp (\w_{y_i} ^\top \h_i)}{ \sum\limits_{c \in [k]} \exp (\w_c^\top \h_i) }  \right) + \w_{y_i}^\top \h_i \right]
\end{equation*}
\endgroup
\end{proposition}

As shown in Prop.~\ref{thm: ce_limit}, in the presence of a large number of prototypes, optimizing SCL is akin to optimizing cross-entropy with a fixed classifier. 
\begin{remark} \label{rem:CE}
    This setting is remarkably similar to \cite{yang2022inducing} that trains CE loss with fixed classifiers forming an ETF geometry. However two key differences emerge: (i) the features and prototypes are normalized, i.e. $\|\h_i\| = 1$ $\forall i \in B$, $\|\w_c\|= 1$ $\forall c \in [k]$, and (ii) here, there is an additional alignment-promoting regularization induced by the inner product between $\h_i$ and $\w_{y_i}$. As we will see below, we also explore  choices of prototypes that deviate from the ETF geomerty.
\end{remark}
As described in Rem.~\ref{rem:CE}, optimizing CE with a fixed classifier has been previously studied \cite{yang2022inducing, hoffer2018fix}; however, typically embeddings are not normalized and different geometries have yet to be considered. In particular, we have empirically found that normalizing embeddings leads to faster geometry convergence consistent with the results of \cite{hoffer2018fix}. Lastly, we arrive at this setting from an entirely different view, one of understanding SCL with prototypes.

Below, we use the simplified loss in Proposition \ref{thm: ce_limit}, to analytically study the geometry of embeddings in the specific setting (that of the experiments of Section \ref{sec:tuning}) where prototypes form an ETF. To facilitate the analysis, we adopt the  unconstrained-features model (UFM) \cite{NC_Mixon}, where the embeddings, $\h_i$, are treated as free variables. 
\begin{proposition}\label{thm: alignment}
    If $\{\w_c\}_{c \in [k]}$ form an equiangular tight frame, i.e. $\w_c ^\top \w_{c'} = \frac{-1}{k-1}$ for $c\neq c'$,  then the optimal embeddings align with their corresponding prototype, $\h_i^* = \w_{y_i}$. 
\end{proposition}
Prop.~\ref{thm: ce_limit} and Prop.~\ref{thm: alignment} (the proofs of which are deferred to the appendix) emphasize the impact of prototypes on SCL optimization showing that in the limit $n_w \gg n$, ETF is the optimal geometry. However, as mentioned in \cite{kini2021labelimbalanced, CDT_Ye,cao2019learning} allowing for better separability for minority classes can potentially improve generalization. Thus we now consider convergence to non-symmetric geometries, which could potentially favor minority separability. 
In Fig.~\ref{fig:heatmap_geom}, we use the limiting form of SCL given in Prop.~\ref{thm: ce_limit} to illustrate the final learnt geometry $\G_{\M}$ of features trained using three possible prototype geometries: 1) \textbf{ETF} which assigns equal angles between prototypes 2) \textbf{Improved Minority Angles} which assigns a larger angle between prototypes belonging to the minority classes and 3) \textbf{Majority Collapse}, an extreme case which assigns the same prototype for all majority classes, forcing the majority class features to collapse to the same vector. Models are trained with a similar setup as in Fig.~\ref{fig:convergence} and Fig.~\ref{fig:heatmap_comp} albeit with learning rate annealing of 0.1 at epochs 200 and 300 as we observed that it expedites convergence. It is clear in Fig.~\ref{fig:heatmap_geom} that the learnt features can be significantly altered based on the choice of prototypes allowing for geometries with more clear separability of minority classes. In summary, these experiments demonstrate the flexibility of SCL with prototypes, and create an opportunity to explore a wide variety of prototype geometries. This exploration could lead to identifying geometries that result in improved test performance when trained under label imbalances. We leave this to future work. 
\begin{Figure}
        \centering
        \includegraphics[width = \textwidth]{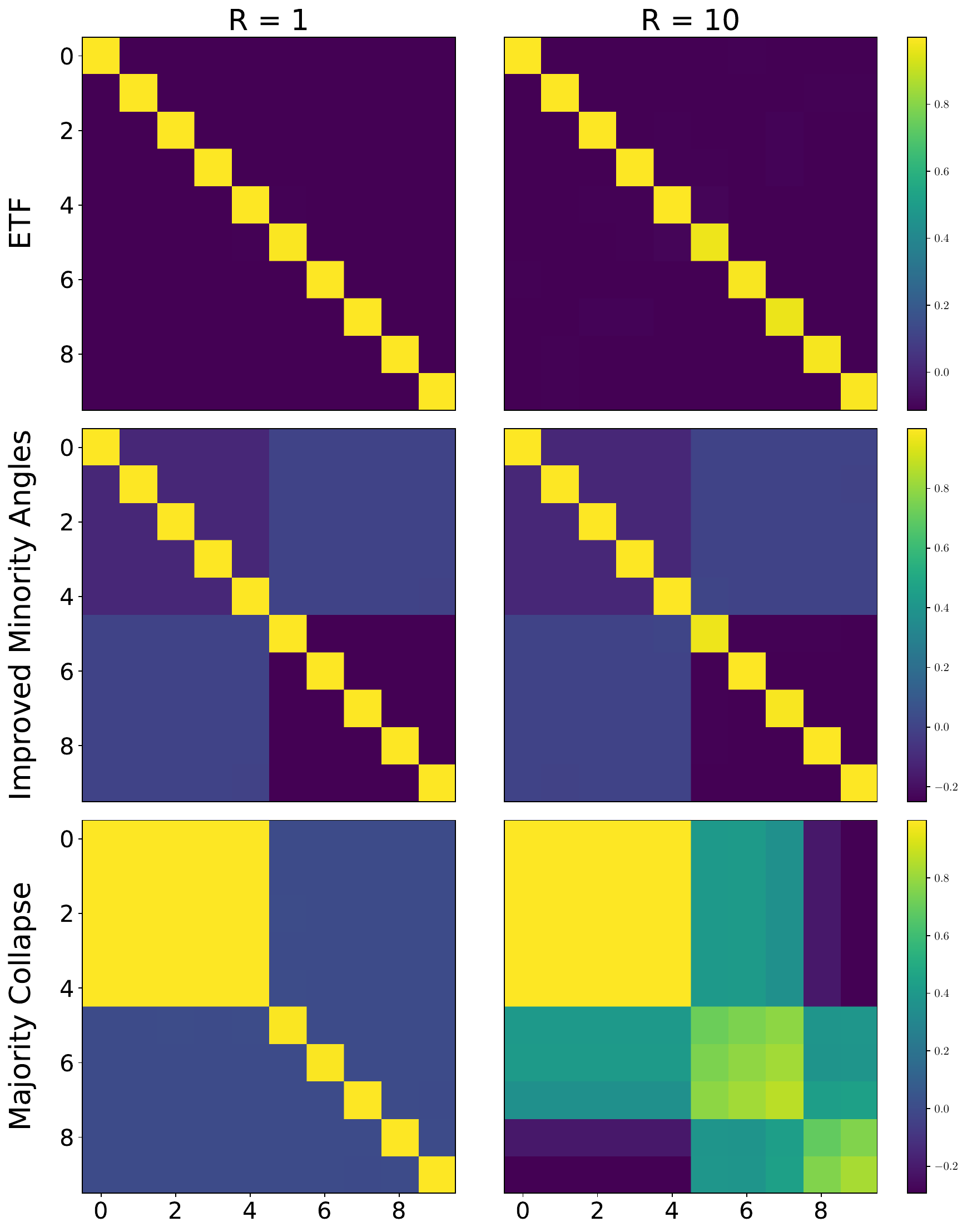}
        \captionof{figure}
        {Comparison of the Gram matrices ($\G_\M$) of learned embeddings with different prototypes trained with the limiting form of SCL given in Thm.~\ref{thm: ce_limit}. (Top) ETF prototypes (Middle) Large Minority Angles (Bottom) Majority Collapse.}
        \label{fig:heatmap_geom} 
\end{Figure}
\vspace{-0.1cm}
\section{Concluding Remarks}
\label{sec:conclusion}
In this work, we have isolated and explored the effects of prototypes on supervised-contrastive loss. In doing so, we have identified a reliable method in tuning the learnt embedding geometry. In addition, a theoretical link to cross-entropy was established.
Overall, our discoveries indicate that employing fixed prototypes offers a promising avenue for streamlining framework modifications that typically treat prototypes as trainable parameters without a clear understanding of their direct contribution. Moreover, this opens up an exciting avenue for future research to explore how choosing prototype geometries favoring larger angles for minority classes can positively impact generalization performance.
\appendix
\section{Proof of Propositions}
\label{sec:majhead}

\subsection{Proof of Proposition \ref{thm: ce_limit}}
\label{ssec:celimProof}
 Let $|B| = n$, and note that $\|\w_c\| = 1, \forall c \in [k]$. Then we have that $\mathcal{L}_B$ with prototypes is of the form, 

\begingroup
\fontsize{9.0pt}{10.0pt} \selectfont
    \begin{equation*}
    \mathcal{L}_B = 
    \sum_{i \in B} \frac{n_w}{n_{B,y_i} + n_w - 1} \mathcal{L}_s
   +\sum_{\hat{c} \in [k]} \frac{n_w}{n_{B,y_i} + n_w - 1} \mathcal{L}_p
\end{equation*}
\endgroup

Here, $\mathcal{L}_s$ is the loss accrued while iterating over each sample in the batch and $\mathcal{L}_p$ is the loss accrued while iterating through each prototype and are given as follows:  

\begingroup
\fontsize{7.0pt}{8.0pt} \selectfont
\begin{multline*}
\mathcal{L}_s :=   \sum_{\substack{j \neq i \\ y_j = y_i}}  \frac{-1}{n_w}\log \left( \frac{\exp (\h_i ^\top \h_j)}{\sum\limits_{\ell \neq i} \exp (\h_\ell^\top \h_i)  + n_w \sum\limits_{c \in [k]} \exp (\w_c^\top \h_i ) }  \right) \\ - 
\log \left( \frac{\exp (\h_i ^\top \w_{y_i})}{\sum\limits_{\ell \neq i} \exp (\h_\ell^\top \h_i)  + 
n_w
\sum\limits_{c \in [k]} \exp (\w_c^\top \h_i ) }  \right)\,,
\end{multline*}
\endgroup

\begingroup
\fontsize{7.0pt}{8.0pt} \selectfont
\begin{multline*}
\mathcal{L}_p := 
  \sum_{\substack{j \in[n] \\ y_j = \hat{c}}} -\log \left( \frac{\exp (\w_{\hat{c}} ^\top \h_j)}{\sum\limits_{\ell \in [n]} \exp (\h_\ell^\top \w_{\hat{c}})  + n_w \Lambda +(n_w -1) e }  \right) \\ - (n_w-1) \log \left( \frac{e}{\sum\limits_{\ell \neq i} \exp (\h_\ell^\top \w_{\hat{c}})  +  n_w \Lambda +(n_w -1) e  }  \right) \,.
\end{multline*}
\endgroup

Above, we have used $\Lambda = \sum_{c \neq \hat{c}} \exp (\w_c^\top \w_{\hat{c}} )$ to denote a fixed constant that is determined after selecting the desired geometry, and for compact notation, we have denote $e=\exp(1)$. 

For clarity, we analyze each term $\mathcal{L}_s$ and $\mathcal{L}_p$ separately. First, as $n_w \gg n_{B,y_i}$ the first term of $\mathcal{L}_s$ is proportional to $\nicefrac{1}{n_w}$, we can neglect it. Moreover, as $n_w$ increases,  $\sum\limits_{\ell \neq i} \exp (\h_\ell^\top \h_i)  \ll n_w \sum\limits_{c \in [k]} \exp (\w_c^\top \h_i ) $. Thus, for large $n_w$ we have that,


\begingroup
\fontsize{9.0pt}{10.0pt} \selectfont
\begin{equation*}
\mathcal{L}_s 
\approx  
-\log \left( \frac{\exp (\h_i ^\top \w_{y_i})}{ n_w \sum\limits_{c \in [k]} \exp (\w_c^\top \h_i ) }  \right) 
\end{equation*}
\endgroup

Now considering $\mathcal{L}_p$, we have that the denominators of the logarithms are approximately given as, 

\begingroup
\fontsize{9.0pt}{10.0pt} \selectfont
\begin{multline*}
\sum\limits_{\ell \in [n]} \exp (\h_\ell^\top \w_{\hat{c}})  + n_w \Lambda +(n_w -1) e \approx n_w \Lambda +(n_w -1) e
\\
\sum\limits_{\ell \neq i} \exp (\h_\ell^\top \w_{\hat{c}})  +  n_w \Lambda +(n_w -1) e  \approx n_w \Lambda +(n_w -1) e
\end{multline*}
\endgroup

Thus we get that, 
\begingroup
\fontsize{9.0pt}{10.0pt} \selectfont
\begin{equation*}
\mathcal{L}_p \approx 
  \sum_{\substack{j \in[n] \\ y_j = \hat{c}}} -\log \left( \frac{\exp (\w_{\hat{c}} ^\top \h_j)}{  n_w \Lambda +(n_w -1) e }  \right) - \Phi \,,
\end{equation*}
\endgroup
where $\Phi := (n_w-1) \log \left( \frac{e}{ n_w \Lambda +(n_w -1) e  }  \right) $. Furthermore, in the limit $\frac{n_w}{n_{B,y_i} + n_w - 1} \rightarrow 1$. 

Combining the above, the per-batch loss ($\mathcal{L}_B$) can be expressed as, 
\begingroup
\fontsize{9.0pt}{10.0pt} \selectfont
    \begin{multline*}
    \mathcal{L}_B \approx 
    \sum_{i \in B} -\log \left( \frac{\exp (\h_i ^\top \w_{y_i})}{ \sum\limits_{c \in [k]} \exp (\w_c^\top \h_i ) }  \right) + \log(n_w)\\
   +\sum_{\hat{c} \in [k]}  \sum_{\substack{j \in[n] \\ y_j = \hat{c}}} -\log \left( \exp (\w_{\hat{c}} ^\top \h_j)   \right) - \Phi + \log(n_w \Lambda +(n_w -1) e )\,.
\end{multline*}
\endgroup

Since the optimal embeddings $\h_i^*$ are independent of any additive constants on the objective it suffices to drop them during optimization. Thus we arrive at the desired:

\begingroup
\fontsize{9.0pt}{10.0pt} \selectfont
    \begin{equation*}
    \mathcal{L}_B \rightarrow
   - \sum_{i \in B} \left[ \log \left( \frac{\exp (\w_{y_i} ^\top \h_i)}{ \sum\limits_{c \in [k]} \exp (\w_c^\top \h_i) }  \right) + \w_{y_i}^\top \h_i \right]\,.
\end{equation*}
\endgroup

\subsection{Proof Sketch of Proposition \ref{thm: alignment}}
\label{ssec:alignmentProof}

    We follow a similar proof technique to \cite{yang2022inducing}, thus we only mention the delicate aspects necessary to handle the alignment term. Consider the minimization program given below, with the objective as given by Prop.~\ref{thm: ce_limit}, while relaxing the norm constraint on the embeddings. 
    \begingroup
\fontsize{9.0pt}{10.0pt} \selectfont
    \begin{equation*}
    \min_{\| \h_i \|^2 \leq 1 } - \sum_{i \in B} \left[ \log \left( \frac{\exp (\w_{y_i} ^\top \h_i)}{ \sum\limits_{c \in [k]} \exp (\w_c^\top \h_i) }  \right) + \w_{y_i}^\top \h_i \right]
\end{equation*}
\endgroup



 Now, as a first step, one can define the Lagrangian $L(\{\h_i\},\{\lambda_i\})$ for $i\in [n]$. Noting that $\{\lambda_i\}_{i \in [n]}$ are the dual variables associated with the norm constraints, as in \cite{yang2022inducing} we can prove by contradiction that $\lambda_i \neq 0$. This implies that $\| \h_i \|^2 = 1$ and $\lambda_i > 0$ from the KKT conditions.
 
 As a next step, we can define $p_y =\frac{\exp (\w_{y} ^\top \h_i)}{ \sum_{c \in [k]} \exp (\w_c^\top \h_i) } $ (as in \cite{yang2022inducing}). From here, one can establish that for $\Tilde{c} \neq \hat{c} \neq y_i$ we have that, 
\begingroup
\fontsize{9.0pt}{10.0pt} \selectfont
 \begin{equation*}
      {\frac{p_{\hat{c}}}{p_{\Tilde{c}} }}= {\frac{\exp(\h_i^\top\w_{\hat{c}})}{\exp(\h_i^\top\w_{\Tilde{c}})}} = \frac{\frac{1}{k-1} - 2 \lambda_i \h_i^\top\w_{\hat{c}}}{\frac{1}{k-1} - 2 \lambda_i \h_i^\top\w_{\Tilde{c}}}
\end{equation*}
\endgroup
Taking $x = \h_i^\top \w_{\hat{c}}$ we define the function $\frac{a - bx}{\exp(x)}$. For CE with fixed ETF classifier, the authors of \cite{yang2022inducing} use the monotonicity of $\frac{\exp(x)}{x}$ to complete the proof. In our case, the function $\frac{a - bx}{\exp(x)}$ is strictly decreasing under the constraints $0\leq a\leq 1 , b>0$ in the interval $x\in [-1, 1]$. Therefore, it holds that $\h_i^\top \w_{\hat{c}} = \h_i^\top \w_{\Tilde{c}} $ and $p_{\hat{c}} = p_{\Tilde{c}} = p$. With this fact established, one can directly take the gradient of the Lagrangian, and solve for $\h_{i}^*$, i.e. set $\nabla_{\h_i}L = 0$. Using the established facts in this proof sketch one will find that $\h_i^* = \w_{y_i}$. 

\bibliographystyle{IEEEbib}
\bibliography{refs}

\end{multicols}
\end{document}